\documentclass[10pt,twocolumn,letterpaper]{article}

\usepackage{iccv}
\usepackage{times}
\usepackage{epsfig}
\usepackage{graphicx}
\usepackage{amsmath}
\usepackage{amssymb}
\usepackage{array}
\usepackage{booktabs}
\usepackage{xcolor}
\usepackage{cite}
\usepackage{multirow}
\usepackage{soul}
\usepackage{subcaption}
\usepackage{algpseudocode}
\usepackage{algorithm}
\usepackage{algorithmicx}
\usepackage{booktabs}

\usepackage[pagebackref=true,breaklinks=true,letterpaper=true,colorlinks,bookmarks=false]{hyperref}

\iccvfinalcopy 

\def\iccvPaperID{XXXX} 

\ificcvfinal\fi
\begin{document}

\title{Temporal Recurrent Networks for Online Action Detection}

\author{
    Mingze Xu$^1$\thanks{The first two authors contributed equally.
    Part of this work was done when MX and MG were interns at Honda Research Institute, USA.}
    \hspace{0.3cm} Mingfei Gao$^2$\footnotemark[1]
    \hspace{0.3cm} Yi-Ting Chen$^3$
    \hspace{0.3cm} Larry S. Davis$^2$
    \hspace{0.3cm} David J. Crandall$^1$ \\[.5ex]
    $^1$Indiana University
    \hspace{0.3cm} $^2$University of Maryland
    \hspace{0.3cm} $^3$Honda Research Institute, USA \\
    {\tt\small \{mx6,djcran\}@indiana.edu, \{mgao,lsd\}@umiacs.umd.edu, ychen@honda-ri.com}
}

\newcommand{\djc}[1]{{\textcolor{blue}{djc says: #1}}}
\newcommand{\mingze}[1]{{\textcolor{brown}{mingze says: #1}}}
\newcommand{\xhdr}[1]{\vspace{5pt} \noindent {\textbf{#1}}}
\definecolor{mypurple}{rgb}{0.851,0.823,0.908}
\definecolor{mygreen}{rgb}{0.812,0.878,0.823}
\definecolor{mypink}{rgb}{0.914,0.824,0.863}
\definecolor{myyellow}{rgb}{0.930,0.851,0.769}
\definecolor{mygray}{rgb}{0.745,0.745,0.745}

\maketitle

\begin{abstract}
    Most work on temporal action detection is formulated as an offline
    problem, in which the start and end times of actions are determined
    after the entire video is fully observed. However, important
    real-time applications including surveillance and
    driver assistance systems
    require identifying actions as soon as each video frame
    arrives, based only on current and historical observations.
    In this paper, we propose a novel framework, Temporal Recurrent
    Network (TRN), to model greater temporal context of a video frame
    by simultaneously performing online action detection and
    anticipation of the immediate future. At each moment in time,
    our approach makes use of both accumulated historical evidence and
    predicted future information to better recognize the action
    that is currently occurring, and integrates both of these
    into a unified end-to-end architecture. We evaluate our approach
    on two popular online action detection datasets, HDD and TVSeries,
    as well as another widely used dataset, THUMOS'14. The results
    show that TRN significantly outperforms the state-of-the-art.
\end{abstract}

\vspace{-5pt}
\section{Introduction}

\begin{figure}[t]
    \begin{center}
        \includegraphics[width=1.0\linewidth]{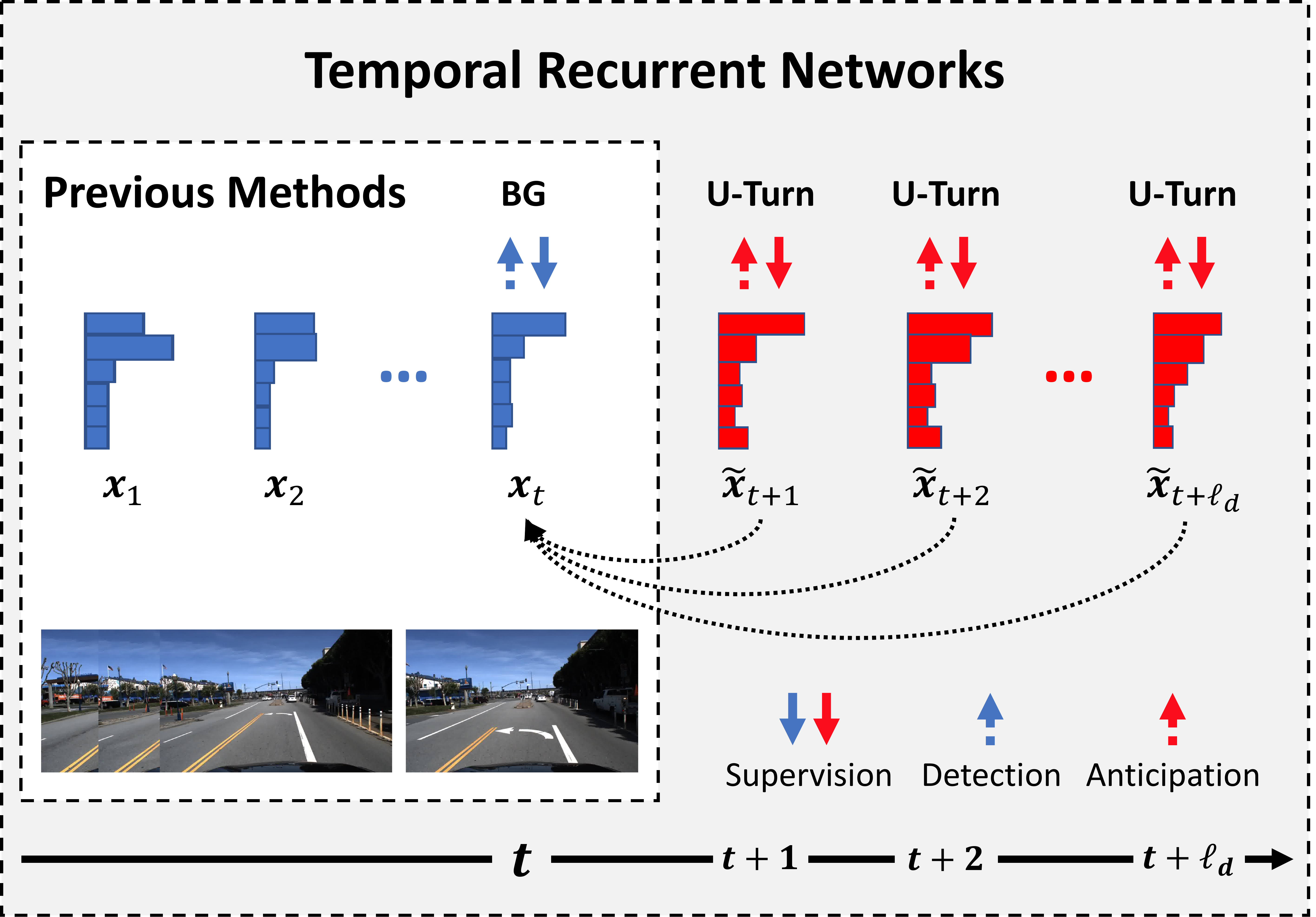}
    \end{center}
    \vspace{-15pt}
    \caption{
        \textit{Comparison between our proposed Temporal Recurrent
        Network (TRN) and previous methods.} Previous methods
        use only historical observations and learn representations for
        actions by optimizing current action estimation. Our approach
        learns a more discriminative representation by jointly
        optimizing current and future action recognition, and
        incorporates the \textit{predicted} future
        information to improve the performance of action detection
        in the present.
    }
    \vspace{-15pt}
    \label{fig:idea}
\end{figure}

As we go about our lives, we continuously monitor the social
environment around us, making inferences about the actions of others
that might affect us. Is that child running into the road or just
walking towards the sidewalk? Is that passerby outstretching his hand
for a punch or a handshake? Is that oncoming car turning left or
doing a U-turn? These and many other actions can occur at any time,
without warning. We must make and update our inferences in real-time
in order to be able to react to the world around us, updating and
refining our hypotheses moment-to-moment as we collect additional
evidence over time.

In contrast, action recognition in computer vision is often studied as
an offline classification problem, in which the goal is to identify a
single action occurring in a short video clip given \textit{all of its
frames}~\cite{shou2016temporal,zhao2017temporal,dai2017temporal,buch2017sst,gao2017turn,chao2018rethinking}.
This offline formulation simplifies the problem considerably:
a left turn can be trivially distinguished from a U-turn if the end of
the action can be observed.
But emerging real-world applications of computer vision like
self-driving cars, interactive home virtual assistants, and
collaborative robots require detecting actions online, in real-time.
Some recent work~\cite{de2016online,gao2017red,shou2018online,de2018modeling}
has studied this problem of online action detection, but the
restriction on using only current and past information makes the
problem much harder.

Here we introduce the novel hypothesis that although future
information is not available in an online setting, \textit{explicitly
predicting the future can help to better classify actions in the
present.} We propose a new model to estimate and use this future
information, and we present experimental results showing that
predicted future information indeed improves the performance of
online action recognition. This may seem like a surprising result
because at test time, a model that predicts the future to infer
an action in the present observes exactly the same evidence
as a model that simply infers the
action directly. However, results in cognitive science and
neuroscience suggest that the human brain uses prediction of the
future as an important mechanism for learning to make estimates
of the present~\cite{n1, n2, n3, n4}.
Our findings seem to confirm that the same applies to automatic
online action recognition, suggesting that 
jointly modeling current action detection and future action
anticipation during training forces the network to learn a more
discriminative representation. 

In more detail, in this paper we propose a general framework called
Temporal Recurrent Network (TRN), in which future information is
predicted as an anticipation task and used together with historical
evidence to recognize action in the current frame
(as shown in Fig.~\ref{fig:idea}). To demonstrate the effectiveness
of our method, we validate TRN on two recent online action detection
datasets (\ie, Honda Research Institute Driving
Dataset (HDD)~\cite{RamanishkaCVPR2018}
and TVSeries~\cite{de2016online}) and a widely used action recognition
dataset, THUMOS'14~\cite{THUMOS14}. Our model is general enough to
use both visual and non-visual sensor data, as we demonstrate for the
HDD driving dataset. Experimental results show that our approach
significantly outperforms baseline methods, especially when
only a fraction of an action is observed. 
We also evaluate action anticipation (predicting the next
action), showing that our method performs better than state-of-the-art
methods even though anticipation is not the focus of this work.

\vspace{-1pt}
\section{Related Work}
\label{sec:related_work}
\vspace{-2pt}

\vspace{-5pt} \xhdr{Action and Activity Recognition.}
There is extensive work in the literature on action and activity
recognition for videos of various types and applications,
from consumer-style~\cite{yue2015beyond} and
surveillance videos~\cite{sultani2018real},
to first-person videos from wearable
cameras~\cite{kitani2011fast,li2015delving,ma2016going}.
Early work used hand-crafted visual features,
such as HOG~\cite{laptev2008learning},
HOF~\cite{laptev2008learning}, and MBH~\cite{wang2013dense},
and motion features, such as improved dense
trajectories~\cite{wang2011action},
while most recent methods use deep convolutional networks. Simonyan
and Zisserman~\cite{simonyan2014two} propose a two-stream
convolutional network that uses both RGB frames and optical flow as
inputs~\cite{wang2016temporal}, while others
including Tran \etal~\cite{tran2015learning} and Carreira
\etal~\cite{carreira2017quo} avoid precomputing
optical flow by learning temporal information
in an end-to-end manner using 3D convolution. Recurrent
neural networks (RNNs), such as long short-term memory
(LSTM)~\cite{hochreiter1997long}
and gated recurrent unit (GRU)~\cite{chung2014empirical} networks
have also been widely adopted to capture temporal
dependencies~\cite{donahue2015long} and motion
information~\cite{extremelylow2018wacv}.
However, most of these methods
focus on trimmed videos and cannot be directly applied to 
long video sequences that contain multiple actions and a wide
diversity of backgrounds.

\xhdr{Offline Action Detection.} Offline methods observe an entire
video and estimate the start and end moment of each action.
Many of these methods are inspired by region-based deep 
networks from object detection~\cite{ren2015faster}
and segmentation~\cite{he2017mask}. Shou~\etal~\cite{shou2016temporal}
propose S-CNNs to localize actions in untrimmed
videos by generating temporal action proposals, and then classifying
them and regressing their temporal boundaries.
TCN~\cite{dai2017temporal} performs proposal ranking but
explicitly incorporates
local context of each proposal. R-C3D~\cite{xu2017r} improves
efficiency by sharing convolutional features 
across proposal generation and classification. 
SST~\cite{buch2017sst}
avoids dividing input videos into overlapping clips, 
introducing more efficient proposal generation in a
single stream, while TURN TAP~\cite{gao2017turn} builds on this
architecture. TAL-Net~\cite{chao2018rethinking} improves
receptive field alignment using a multi-scale architecture that
better exploits temporal context
for both proposal generation and action classification.
CDC~\cite{shou2017cdc} makes frame-level dense predictions
by simultaneously performing spatial downsampling
and temporal upsampling operations.
But the above work assumes all video frames
can be observed, which is not possible in the online task
that we consider here.

\begin{figure*}[t]
    \begin{center}
        \includegraphics[width=1.0\linewidth]{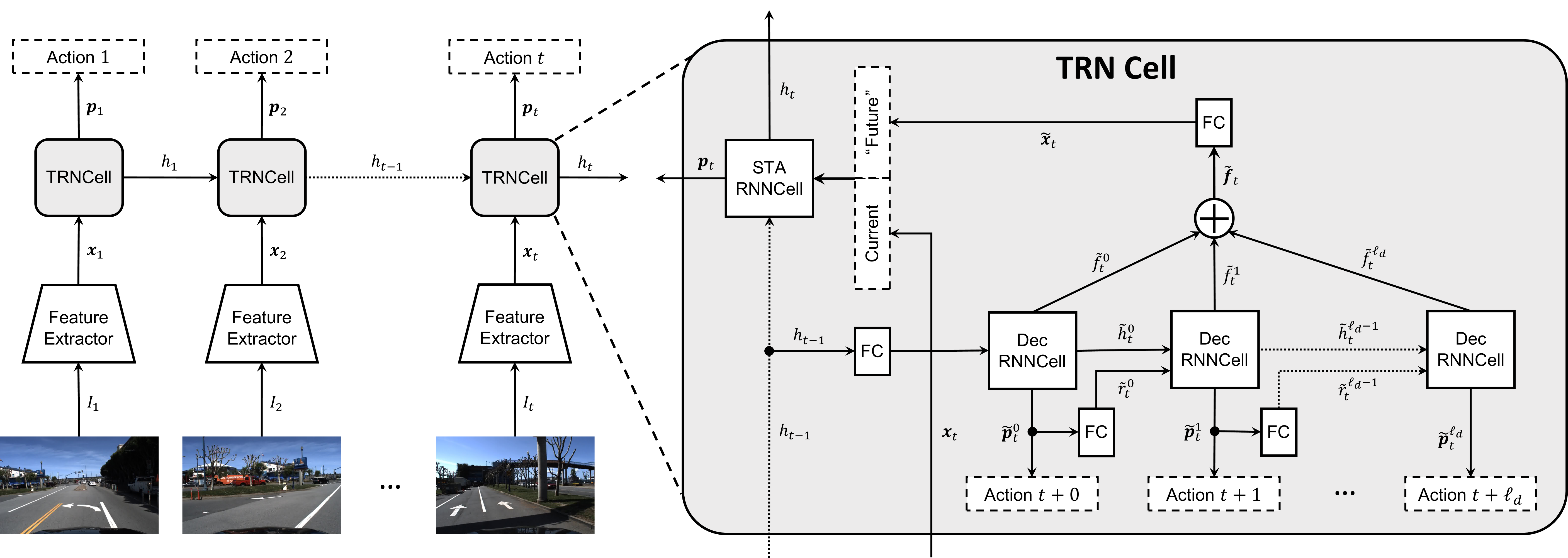}
    \end{center}
    \vspace{-15pt}
    \caption{
        {\textit{Our proposed Temporal Recurrent Network (TRN)},
        which sequentially processes input video frames and outputs
        frame-level action class probabilities, 
        like any RNN. But while RNNs only model historical temporal
        dependencies, TRN anticipates the future via a temporal
        decoder, and incorporates that predicted information to
        improve online action detection.}
    }
    \label{fig:network}
\end{figure*}

\xhdr{Early Action Detection.}
Our work is also related to early action detection,
which tries to recognize actions after observing only a fraction of
the event. Hoai~\etal~\cite{hoai2014max} propose a max-margin framework
using structured SVMs for this problem. Ma~\etal~\cite{ma2016learning}
design an improved
technique based on LSTMs and modify the training loss
based on an assumption that the score margin between the
correct and incorrect classes should be non-decreasing as
more observations are made.

\xhdr{Online Action Detection.}
Given a live video stream, online action detection tries to detect
the actions performed in each frame as soon as it arrives, without
considering future context. De Geest~\etal~\cite{de2016online}
first presented a concrete definition
and realistic dataset (TVSeries) for this problem.
They later~\cite{de2018modeling}
proposed a two-stream feedback network, with one stream focusing
on input feature interpretation and the other modeling temporal
dependencies between actions. Gao~\etal~\cite{gao2017red} propose
a Reinforced Encoder-Decoder (RED) network 
and a reinforcement loss to encourage 
recognizing  actions as early as possible. RED was designed
for action anticipation
-- predicting actions a few seconds into
the future -- but can be applied to online detection
by setting the anticipation time to $0$.
Shou~\etal~\cite{shou2018online}
identify the start time of each action 
using Generative Adversarial Networks and adaptive
sampling to distinguish ambiguous backgrounds,
and explicit temporal modeling around transitions
between actions for temporal consistency.

In contrast to all of this existing online action detection work which
only focuses on current and past observations, we 
introduce a model that learns to simultaneously perform online
action detection and anticipation
of the immediate future, and uses this estimated
``future information'' to improve the action
detection performance of the present.

\section{Online Action Detection}

Given a live video stream that contains one or more actions,
our goal is to recognize actions of interest occurring in each video
frame. Unlike most prior work that assumes the entire video is
available at once, this \textit{online action detection} problem
requires us to process
each frame as soon as it arrives, without accessing
\textit{any} future information. 
More formally, our goal is to estimate, for each frame $I_t$ of
an image sequence, a probability distribution
$\textbf{p}_t = [\,p_t^0, p_t^1, p_t^2, \cdots, p_t^K]$
over $K$ possible actions, given only the past and current frames,
$\{I_{1}, I_{2}, \cdots, I_{t}\}$ (where
$p_t^0$ denotes the ``background'' probability
that no action is occurring).

\subsection{Temporal Recurrent Network (TRN)}

To solve this problem, we introduce a novel framework called Temporal
Recurrent Network (TRN). The main idea is to train
a network that predicts actions several frames into the future,
and then uses that prediction to classify an action in the present.
Fig.~\ref{fig:network} shows the architecture of TRN.  
The core of the network is a
powerful recurrent unit, the \textit{TRN cell}. Like a general RNN
cell, at each time $t$ a TRN cell receives a
feature vector $\textbf{x}_t$
corresponding to the observation at time $t$,
which could include 
some combination of evidence from the appearance or motion in
frame $I_t$ or even other sensor modalities collected at time $t$,
and the hidden state $h_{t-1}$ from the previous time
step. The cell then outputs $\textbf{p}_t$, a probability distribution
estimating which action is happening in $I_t$. 
The hidden state $h_t$ is then updated
and used for estimating the next time step. But while a traditional
RNN cell only models the \textit{prior} temporal dependencies by
accumulating historical evidence of the input sequence, a TRN cell
also takes advantage of the temporal correlations between current and
future actions by anticipating upcoming actions and explicitly
utilizing these estimates to help recognize the present action.

\subsection{TRN Cell}
The TRN cell controls the flow of  internal information by using
a temporal decoder, a future gate,
and a spatiotemporal accumulator (STA).
We use LSTMs~\cite{hochreiter1997long} as the backbone for both the
temporal decoder and the STA in our implementation, although other
temporal models such as gated recurrent units
(GRUs)~\cite{chung2014empirical}
and temporal convolutional networks (TCNs)~\cite{lea2017temporal}
could be used.
The temporal decoder learns a feature representation and predicts
actions for the future sequence.
The future gate receives a vector of hidden states from the decoder
and embeds these features as the future context.
The STA captures the spatiotemporal features from historical,
current, and predicted future information, and estimates the action
occurring in the current frame.

\begin{algorithm}
    \caption{Workflow of TRN Cell}
    \hspace*{\algorithmicindent} \textbf{Input}: image feature $\textbf{x}_t$ and previous hidden state $h_{t-1}$ \\
    \hspace*{\algorithmicindent} \textbf{Output}: probabilities $\textbf{p}_t$ and current hidden state $h_{t}$
    \begin{algorithmic}[1]
        \State Initialize $\widetilde{h}_t^{-1}$ with $h_{t-1}$ embedded by an FC layer
        \State Initialize $\widetilde{r}_t^{-1}$ with all zeros
        \For{i = 0 : $l_d$}
        \State Update $\widetilde{h}_t^i$ using $\widetilde{r}_t^{i-1}$ and $\widetilde{h}_t^{i-1}$
        \State Compute $f_t^i$ and $\widetilde{p}_t^i$ using $\widetilde{h}_t^i$
        \State Update $\widetilde{r}_t^{i}$ using $\widetilde{p}_t^i$
        \EndFor
        \State Compute future context features $\widetilde{\textbf{x}}_t$ as Eq. (\ref{eq:avgpool})
        \State Update $h_t$ with $\text{STA}(h_{t-1}, [\textbf{x}_t, \widetilde{\textbf{x}}_t])$
        \State Compute $\textbf{p}_t$ as Eq. (\ref{eq:classify})
    \end{algorithmic}
    \label{alg:trncell}
\end{algorithm}

We now describe each component of a TRN cell in detail,
as summarized in Alg.~\ref{alg:trncell}.

\xhdr{The temporal decoder} works sequentially to output the estimates
of future actions and their corresponding hidden states
$\{\widetilde{h}_t^0, \widetilde{h}_t^1, \cdots, \widetilde{h}_t^{\ell_d}\}$
for the next $\ell_d$ time steps, where 
$h_t^i$ for $i \in [0, \ell_d]$
indicates the hidden state at the $i$-th time step after  $t$. The input to the
decoder at the first time step is all zeros. At other time steps $t$,
we feed in the predicted action scores $\widetilde{r}_t^{i-1}$,
embedded by a linear transformer.

\xhdr{The future gate} takes  hidden states from the decoder and
models the feature representation of future context. For simplicity, our
default future gate is an average pooling operator followed by an
fully-connected (FC) layer, but other fusion operations such as non-local (NL)
blocks~\cite{wang2018non} could be used. More formally, the
future context feature $\widetilde{\textbf{x}}_t$ is obtained by
averaging and embedding the hidden state vector, $\widetilde{\textbf{h}}_t$,
gathered from all decoder steps,
\begin{equation}
    \widetilde{\textbf{x}}_t = \mbox{ReLU}(\textbf{W}_f^T \mbox{AvgPool} (\widetilde{\textbf{h}}_t) +\textbf{b}_f).
    \label{eq:avgpool}
\end{equation}
\noindent \textbf{The spatiotemporal accumulator (STA)} takes  the
previous hidden state
$h_{t-1}$ as well as the concatenation of the image feature $\textbf{x}_t$
extracted from $I_t$ and the predicted future feature
$\widetilde{\textbf{x}}_t$ from the future gate, and updates
its hidden states $h_t$. It then calculates a distribution
over candidate actions,
\begin{equation}
    \textbf{p}_t = \mbox{softmax}(\textbf{W}_c^T h_t+\textbf{b}_c),
    \label{eq:classify}
\end{equation}
where $\textbf{W}_c$ and $\textbf{b}_c$ are the parameters of the
FC layer used for action classification.

As we can see, in addition to the estimated action of the current
frame $t$, TRN outputs predicted actions for the next $\ell_d$
time steps. In order to ensure a good future
representation and jointly optimize online action detection
and prediction, we combine the accumulator and decoder losses
during training, \ie the loss of one input sequence is
\begin{equation}
    \sum_{t}\big(loss(\textbf{p}_t, l_t)+ \alpha
    \sum_{i=0}^{\ell_d}loss(\widetilde{\textbf{p}}_{t}^{i}, l_{t+i})\big),
    \label{eq:loss}
\end{equation}
where $\widetilde{\textbf{p}}_{t}^{i}$ indicates the action probabilities
predicted by the decoder for step $i$ after time $t$, $l_t$ represents
the ground truth, $loss$ denotes cross-entropy loss, and
$\alpha$ is a scale factor. We optimize the network using 
offline training  in which labels of both current and future
frames are used. At test time, our
model uses the \textit{predicted} future information \textit{without
accessing actual future frames}, and thus is an online model.

\section{Experiments}

We evaluated our online action detector against multiple
state-of-the-art and baseline methods on three publicly-available
datasets: HDD~\cite{RamanishkaCVPR2018}, TVSeries~\cite{de2016online},
and THUMOS'14~\cite{THUMOS14}. 
We chose these datasets because they
include long, untrimmed videos from diverse
perspectives and applications:
HDD consists of on-road driving from a first-person
(egocentric) view recorded by a front-facing dashboard camera, TVSeries
was recorded from television and contains a variety of
everyday activities, and THUMOS'14 is a popular dataset of
sports-related actions.

\subsection{Datasets}

\vspace{-5pt} \xhdr{HDD}\cite{RamanishkaCVPR2018} includes nearly 104
hours of 137 driving sessions in the San Francisco Bay Area. The
dataset was collected from a vehicle with a front-facing
camera, and includes frame-level annotations of 11 goal-oriented
actions (\eg, intersection passing, left turn,
right turn, \etc). The dataset also includes readings from a
variety of non-visual sensors collected by the instrumented vehicle's
Controller Area Network (CAN bus). We followed prior
work~\cite{RamanishkaCVPR2018} and used 100 sessions for training and
37 sessions for testing.

\xhdr{TVSeries}~\cite{de2016online} contains 27 episodes of 6 popular
TV series, totaling 16 hours of video. The dataset is temporally
annotated at the frame level with 30 realistic, everyday actions (\eg,
pick up, open door, drink, \etc). The dataset is challenging with
diverse actions, multiple actors,
unconstrained viewpoints, heavy occlusions, and a large proportion of
non-action frames.

\xhdr{THUMOS'14}~\cite{THUMOS14} includes over 20 hours of video
of sports annotated with 20 actions. The training set contains only trimmed videos
that cannot be used to train temporal action detection models, so
we followed prior work~\cite{gao2017red}
and train on the validation set (200 untrimmed videos) and evaluate on the test set (213 untrimmed videos).

\subsection{Implementation Details}
We implemented our proposed Temporal Recurrent Network (TRN) in 
PyTorch~\cite{pytorch}, and performed all
experiments on a system with Nvidia Quadro P6000 graphics
cards.\footnote{The code will be made publicly available upon
publication.} To learn the network weights, we used the
Adam~\cite{kingma2014adam} optimizer with default parameters, learning
rate $0.0005$, and weight decay $0.0005$. For data augmentation,
we randomly chopped off $\Delta \in [1, \ell_e]$ frames from the 
beginning for \textit{each} epoch, and discretized the
video of length $L$ into $(L-\Delta)/\ell_e$ non-overlapping training
samples, each with $\ell_e$ consecutive frames.
Our models were optimized in an end-to-end manner using a batch size
of $32$, each with $\ell_e$ input sequence length.
The constant $\alpha$ in Eq.~(\ref{eq:loss})
was set to $1.0$.

\subsection{Settings}
To permit fair comparisons with the
state-of-the-art~\cite{RamanishkaCVPR2018,de2016online,gao2017red},
we follow their experimental settings, including input features and
hyperparameters.

\xhdr{HDD.}
We use the same setting as in~\cite{RamanishkaCVPR2018}.
Video frames and values from CAN bus sensors are first sampled at
$3$ frames per second (fps). The outputs of the \texttt{Conv2d\_7b\_1x1} layer in
InceptionResnet-V2~\cite{szegedy2016inception} pretrained on
ImageNet~\cite{deng2009imagenet} are extracted as the visual
feature for each frame. To preserve spatial information, we
apply an additional $1\times1$ convolution to reduce the extracted
frame features from $8\times8\times1536$ to
$8\times8\times20$, and flatten them into $1200$-dimensional vectors.
Raw sensor values are passed into a fully-connected layer with
$20$-dimensional outputs. These visual and sensor features are then
concatenated as a \textit{multimodal} representation for each video
frame.  We follow~\cite{RamanishkaCVPR2018} and set the input sequence
length $\ell_e$ to  $90$. The number of decoder steps $\ell_d$ is
treated as a hyperparameter that we cross-validate in 
experiments. The hidden units of both the temporal decoder and the STA
are set to $2000$ dimensions.

\xhdr{TVSeries and THUMOS'14.}
We use the same setting as in~\cite{gao2017red}. We extract video
frames at $24$ fps and set the video chunk size to $6$. Decisions are
made at the chunk level, and thus performance is evaluated every $0.25$
seconds. We use two different feature extractors,
VGG-16~\cite{simonyan2014very} and two-stream (TS)
CNN~\cite{xiong2016cuhk}. VGG-16 features are extracted at the
\texttt{fc6} layer from the central frame of each chunk. For the
two-stream features, the appearance features are extracted at the
\texttt{Flatten\_673} layer of ResNet-200~\cite{he2016deep} from the
central frame of each chunk, and the motion features are extracted at
the \texttt{global\_pool} layer of BN-Inception~\cite{ioffe2015batch}
from precomputed optical flow fields between  $6$ consecutive
frames. The appearance and motion features are then concatenated to
construct the two-stream features. The input sequence length $\ell_e$
is set to  $64$ due to GPU memory limitations. Following the
state-of-the-art~\cite{gao2017red}, the number of decoder steps
$\ell_d$ is set to $8$, corresponding to $2$ seconds.
As with HDD, our experiments report results with
different decoder steps. The hidden units of both the temporal decoder
and the STA are set to $4096$ dimensions.

\begin{table*}[t]
    \small
    \centering
    \resizebox{\textwidth}{!}{
        \begin{tabular}
            {@{\quad}lc@{\;}c@{\;}@{\;}c@{\;}@{\;}c@{\;}@{\;}c@{\;}@{\;}c@{\;}@{\;}c@{\;}@{\;}c@{\;}@{\;}c@{\;}@{\;}c@{\;}@{\;}c@{\;}@{\;}c@{\;}@{\quad\;\;\;\;}c@{\quad}}
\toprule
& & \multicolumn{11}{c}{Individual actions} & \\
\cmidrule(r{4\cmidrulekern}){3-13}
            \begin{tabular}{@{}c@{}} \\ Method \end{tabular} &
            \begin{tabular}{@{}c@{}} \\ Inputs \end{tabular} &
            \begin{tabular}{@{}c@{}}intersection \\ passing \end{tabular} &
            \begin{tabular}{@{}c@{}}\\ L  turn            \end{tabular} &
            \begin{tabular}{@{}c@{}}\\ R  turn           \end{tabular} &
            \begin{tabular}{@{}c@{}}L  lane \\ change  \end{tabular} &
            \begin{tabular}{@{}c@{}} R  lane \\ change \end{tabular} &
            \begin{tabular}{@{}c@{}} L  lane \\ branch  \end{tabular} &
            \begin{tabular}{@{}c@{}} R  lane \\ branch \end{tabular} &
            \begin{tabular}{@{}c@{}}crosswalk \\ passing    \end{tabular} &
            \begin{tabular}{@{}c@{}}railroad \\ passing     \end{tabular} &
            \begin{tabular}{@{}c@{}} \\merge                   \end{tabular} &
            \begin{tabular}{@{}c@{}} \\u-turn                  \end{tabular} &
            \begin{tabular}{@{}c@{}}Overall \\mAP                  \end{tabular} \\
            \midrule
            CNN & \multirow{4}{*}{Sensors} & 34.2 & 72.0 & 74.9 & 16.0 &  8.5 & 7.6 & 1.2 & 0.4 & 0.1 & 2.5 & 32.5 & 22.7 \\
            LSTM~\cite{RamanishkaCVPR2018} & & 36.4 & 66.2 & 74.2 & 26.1 & 13.3 & 8.0 & 0.2 & 0.3 & 0.0 & 3.5 & 33.5 & 23.8 \\
            ED & & 43.9 & 73.9 & 75.7 & 31.8 & 15.2 & 15.1 & 2.1 & 0.5 & 0.1 & 4.1 & 39.1 & 27.4 \\
            \textbf{TRN} & & 46.5 & 75.2 & 77.7 & 35.9 & 19.7 & 18.5 & 3.8 & 0.7 & 0.1 & 2.5 & 40.3 &\textbf{29.2} \\
            \cmidrule{1-14}
            CNN & \multirow{4}{*}{InceptionResNet-V2} & 53.4 & 47.3 & 39.4 & 23.8 & 17.9 & 25.2 & 2.9 & 4.8 & 1.6 & 4.3 & 7.2 & 20.7 \\
            LSTM~\cite{RamanishkaCVPR2018} & & 65.7 & 57.7 & 54.4 & 27.8 & 26.1 & 25.7 & 1.7 & 16.0 & 2.5 & 4.8 & 13.6 & 26.9 \\
            ED & & 63.1 & 54.2 & 55.1 &28.3 & 35.9 & 27.6 & 8.5 & 7.1 & 0.3 & 4.2 & 14.6 & 27.2 \\
            \textbf{TRN} & & 63.5 & 57.0 & 57.3 & 28.4 & 37.8 & 31.8 & 10.5 & 11.0 & 0.5 & 3.5 & 25.4 &\textbf{29.7} \\
            \cmidrule{1-14}
            CNN & \multirow{4}{*}{Multimodal} & 73.7 & 73.2 & 73.3 & 25.7 & 24.0 & 27.6 & 4.2 & 4.0 & 2.8 & 4.7 & 30.6 & 31.3 \\
            LSTM~\cite{RamanishkaCVPR2018} & & 76.6 & 76.1 & 77.4 & 41.9 & 23.0 & 25.4 &  1.0 & 11.8 & 3.3 & 4.9 & 17.6 & 32.7 \\
            ED & & 77.2 & 74.0 & 77.1 & 44.6 & 41.4 & 36.6 & 4.1 & 11.4 & 2.2 & 5.1 & 43.1 & 37.8 \\
            \textbf{TRN} & & 79.0 & 77.0 & 76.6 & 45.9 & 43.6 & 46.9 & 7.5 & 13.4 & 4.5 & 5.8 & 49.6 &\textbf{40.8} \\
\bottomrule
        \end{tabular}
    }
    \vspace{-6pt}
    \caption{\textit{Results of online action detection
    on HDD,} comparing TRN and baselines using 
    mAP (\%).}
    \vspace{-6pt}
    \label{table:hdd_detection}
\end{table*}


\subsection{Evaluation Protocols}
\label{sec:protocols}

We follow most existing work and use per-frame \textbf{mean average
precision (mAP)} to evaluate the performance of online action
detection. We also use per-frame \textbf{calibrated average
precision (cAP)}, which was proposed in~\cite{de2016online} to better
evaluate online action detection on TVSeries,
\begin{equation}
    cAP = \frac{\sum_{k}cPrec(k)*I(k)}{P},
    \label{eq:cap}
\end{equation}
where calibrated precision $cPrec=\frac{TP}{TP+FP/w}$,
$I(k)$ is $1$ if frame $k$ is a true
positive, $P$ denotes the total number of true positives, and $w$ is
the ratio between negative and positive frames. The advantage of cAP
is that it corrects for class imbalance between positive
and negative samples.

Another important goal of online
action detection is to recognize actions as early as
possible; \ie, an approach should be rewarded if it
produces high scores for target actions at their early stages (the
earlier the better).
To investigate our performance at different time stages, we
follow~\cite{de2016online} and compute mAP or cAP for each decile
(ten-percent interval) of the video frames separately. For example,
we compute mAP or cAP on the first $10\%$ of the frames of the action,
then the next $10\%$, and so on.

\begin{table}[t]
    \centering
    \small
    \begin{tabular}
        {@{\quad}l@{\quad}@{\quad}c@{\quad}@{\quad}c@{\quad}}
        \toprule
        Method & Inputs & mcAP \\
        \midrule
        {CNN}~\cite{de2016online} & \multirow{7}{*}{VGG} & 60.8 \\
        {LSTM}~\cite{de2016online} & & 64.1 \\
        {RED}~\cite{gao2017red} & & 71.2 \\
        {Stacked LSTM}~\cite{de2018modeling} & & 71.4 \\
        {2S-FN}~\cite{de2018modeling} & & 72.4\\
        \textbf{TRN} & &\textbf{75.4} \\
        \midrule
        {SVM}~\cite{de2016online} & FV & 74.3 \\
        \midrule
        {RED}~\cite{gao2017red} & \multirow{2}{*}{TS} & 79.2 \\
        \textbf{TRN} & &\textbf{83.7} \\
        \bottomrule
    \end{tabular}
    \vspace{-5pt}
    \caption{\textit{Results of online action detection on TVSeries,}
    comparing TRN and the state-of-the-art using 
    cAP (\%).
    }
    \vspace{-5pt}
    \label{table:tv_detection}
\end{table}

\begin{table}[t]
    \centering
    \small
    \begin{tabular}
        {@{\quad}l@{\quad}@{\quad\;\;\;\;\;\;\;\;\;\;\;}c@{\quad}}
        \toprule
        Method & mAP \\
        \midrule
        {Single-frame CNN}~\cite{simonyan2014very} &34.7 \\
        {Two-stream CNN}~\cite{simonyan2014two} &36.2 \\
        {C3D + LinearInterp}~\cite{shou2017cdc} &37.0 \\
        {Predictive-corrective}~\cite{dave2017predictive} &38.9 \\
        {LSTM}~\cite{donahue2015long} &39.3 \\
        {MultiLSTM}~\cite{yeung2018every} &41.3\\
        {Conv \& De-conv}~\cite{shou2017cdc} &41.7\\
        {CDC}~\cite{shou2017cdc} &44.4 \\
        {RED~\cite{gao2017red}} &45.3 \\
        \midrule
        \textbf{TRN} &\textbf{47.2} \\
        \bottomrule
    \end{tabular}
    \vspace{-5pt}
    \caption{\textit{Results of online action detection on THUMOS'14,}
    comparing TRN and the state-of-the-art using mAP (\%).}
    \vspace{-10pt}
    \label{table:th_detection}
\end{table}


\subsection{Baselines}

We compared against multiple baselines to confirm the effectiveness
of our approach.

\xhdr{CNN} baseline models~\cite{simonyan2014very,simonyan2014two} consider online action detection as a
general image classification problem. These
baselines identify the action
in each individual video frame without modeling
temporal information.
For TVSeries and THUMOS'14, we reprint
the results of
CNN-based methods from De Geest \etal~\cite{de2016online}
and Shou \etal~\cite{shou2017cdc}. For HDD, we follow
Ramanishka \etal~\cite{RamanishkaCVPR2018} and use
InceptionResnet-V2~\cite{szegedy2016inception} pretrained on ImageNet
as the backbone and finetune the last fully-connected layer
with softmax to estimate class probabilities.

\xhdr{LSTM} and variants have been widely
used in action detection~\cite{RamanishkaCVPR2018,yeung2018every}.
LSTM networks model the dependencies between consecutive 
frames and jointly
capture spatial and temporal information of the video sequence. For
each frame, the LSTM receives the image features and the previous
hidden state as inputs, and outputs a probability distribution over
candidate actions.

\xhdr{Encoder-Decoder (ED)} architectures~\cite{cho2014learning}
also model temporal dependencies.
The encoder is similar to a general LSTM and summarizes  historical visual information into a feature vector.
The decoder is also an LSTM that produces
predicted representations for the future sequence based only on these
encoded features. Since there are no published results of ED-based
methods on HDD, we implemented a baseline with the same experimental
settings as TRN, including input features, hyperparameters,
loss function, \etc

\xhdr{Stronger Baselines.}
In addition to the above basic baselines, we tested three
types of stronger baselines that were designed for online action 
detection on TVSeries and THUMOS'14.
\textbf{Convolutional-De-Convolutional (CDC)}~\cite{shou2017cdc}
places CDC filters on top of a 3D CNN and integrates two reverse
operations, spatial downsampling and temporal upsampling, to
precisely predict actions at a frame-level.
\textbf{Two-Stream Feedback Network (2S-FN)}~\cite{de2018modeling}
is built on an LSTM with two recurrent units, where one stream focuses
on the input interpretation and the other models temporal dependencies
between actions.
\textbf{Reinforced Encoder-Decoder (RED)}~\cite{gao2017red} with
a dedicated reinforcement loss is an advanced version of ED,
and currently performs the best among all the baselines for online
action detection.

\subsection{Results}

\subsubsection{Evaluation of Online Action Detection}
\vspace{-5pt} Table~\ref{table:hdd_detection} presents evaluation
results on HDD dataset. TRN significantly outperforms the
state-of-the-art, Ramanishka \etal~\cite{RamanishkaCVPR2018}, by
$5.4\%$, $2.8\%$, and $8.1\%$ in terms of mAP with sensor data,
InceptionResnet-v2, and multimodal features as inputs, respectively.
Interestingly, the performance gaps between TRN
and~\cite{RamanishkaCVPR2018} are much
larger when the input contains sensor data. Driving
behaviors are highly related to CAN bus signals, such as steering
angle, yaw rate, velocity, \etc, and this result suggests that TRN can better take
advantage of these useful input cues.
Table~\ref{table:tv_detection} presents comparisons between TRN and
baselines on TVSeries. TRN
significantly outperforms the state-of-the-art using VGG (mcAP of $3.0\%$ over
2S-FN~\cite{de2018modeling})  and 
two-stream input features (mcAP of $4.5\%$ over
RED~\cite{gao2017red}). We also
evaluated TRN on THUMOS'14 in
Table~\ref{table:th_detection}. The results show that TRN outperforms
all the baseline models (mAP of $1.9\%$ over RED~\cite{gao2017red} and
$2.8\%$ over CDC~\cite{shou2017cdc}).

\begin{figure*}
    \center
    \begin{subfigure}[t]{1.0\textwidth}
        \center
        \includegraphics[width=0.99\linewidth]{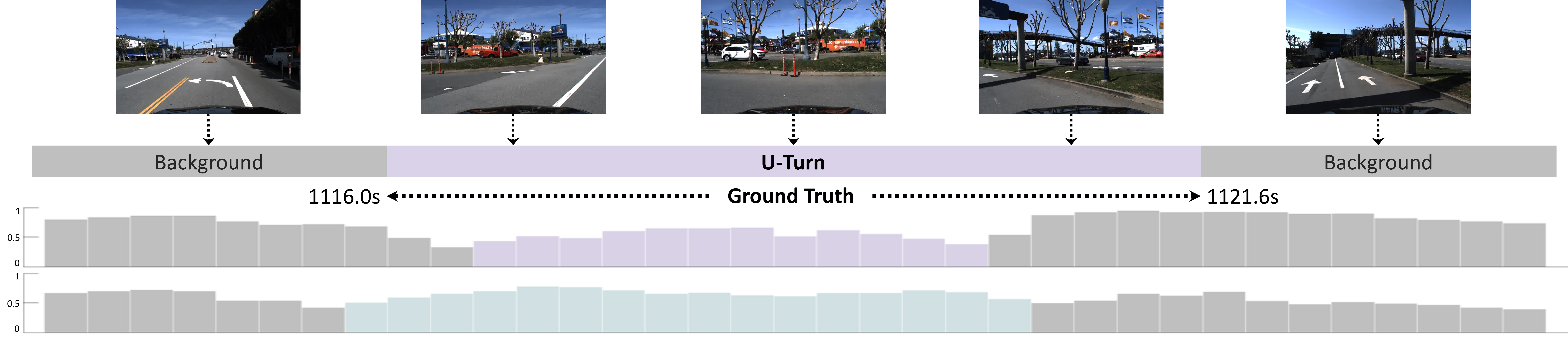}
        \vspace{-6pt}
        \caption{
            Qualitative comparison between our approach (3rd row)
            and~\cite{RamanishkaCVPR2018} (4th row) on HDD dataset.
            \textit{U-Turn} is shown in \textbf{\textcolor{mypurple}{purple}},
            \textit{Left Turn} is shown in \textbf{\textcolor{mygreen}{green}},
            and \textit{Background} is shown in \textbf{\textcolor{mygray}{gray}}.
        }
        \vspace{4pt}
        \label{fig:exp_hdd}
    \end{subfigure}
    \begin{subfigure}[t]{1.0\textwidth}
        \center
        \includegraphics[width=0.99\linewidth]{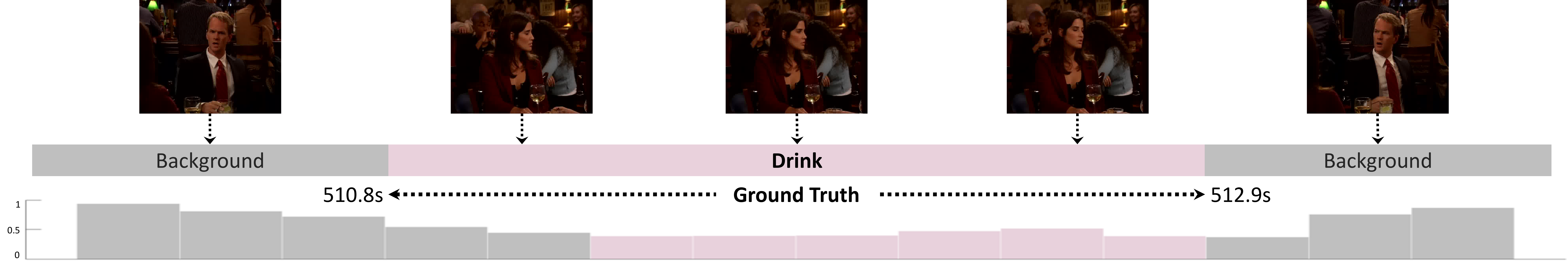}
        \vspace{-6pt}
        \caption{
            Qualitative result of our approach (3rd row) on TVSeries
            dataset. \textit{Drink} is shown in \textbf{\textcolor{mypink}{pink}} and
            \textit{Background} is shown in \textbf{\textcolor{mygray}{gray}}.
        }
        \vspace{4pt}
        \label{fig:exp_tv}
    \end{subfigure}
    \begin{subfigure}[t]{1.0\textwidth}
        \center
        \includegraphics[width=0.99\linewidth]{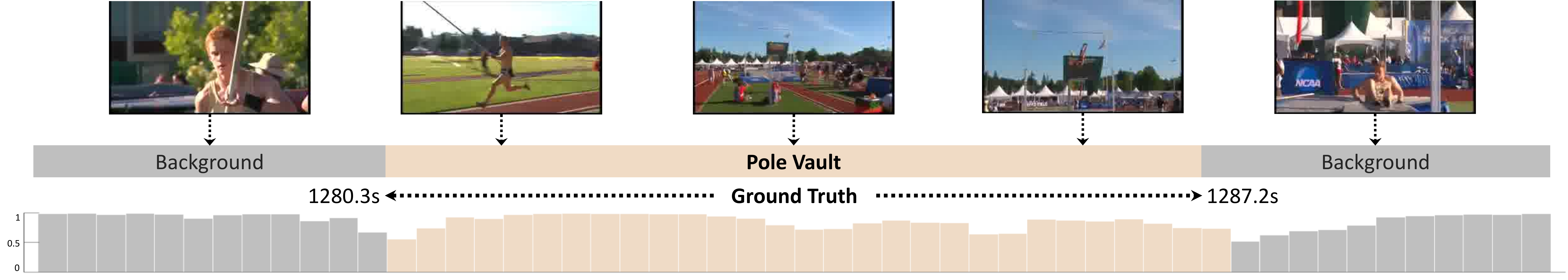}
        \vspace{-6pt}
        \caption{
            Qualitative result of our approach (3rd row) on THUMOS'14
            dataset. \textit{Pole Vault} is shown in \textbf{\textcolor{myyellow}{yellow}}
            and \textit{Background} is shown in \textbf{\textcolor{mygray}{gray}}.
        }
        \vspace{4pt}
        \label{fig:exp_th}
    \end{subfigure}
    \vspace{-10pt}
    \caption{
        \textit{Qualitative results of our approach and baselines on HDD, TVSeries, and
        THUMOS'14 datasets.} 
The vertical bars
        indicate the scores of the predicted class. (Best viewed in color.) 
    }
    \label{fig:results}
    \vspace{-10pt}
\end{figure*}

\xhdr{Qualitative Results} are shown in Fig.~\ref{fig:results}.
In Fig.~\ref{fig:exp_hdd}, we visualize and compare the results of TRN,
and compare with Ramanishka \etal~\cite{RamanishkaCVPR2018} on HDD. As shown,
\textit{u-turn} is difficult to classify from a first-person perspective
because the early stage is nearly indistinguishable from
\textit{left turn}. With the help of the learned better representation
and predicted future information, TRN differentiates from
subtle differences and ``look ahead'' to reduce this
ambiguity. As shown in Table~\ref{table:hdd_detection}, TRN beats the
baseline models on most of the actions using \textit{multimodal}
inputs, but much more significantly on these ``difficult'' classes, such as
\textit{lane branch} and \textit{u-turn}.
Qualitative
results also clearly demonstrate that TRN produces not only the correct
action label, but also better boundaries. Fig.~\ref{fig:exp_tv}
and~\ref{fig:exp_th} show promising results on
TVSeries and THUMOS'14. Note that TVSeries is very
challenging; for example, the drinking action in Fig.~\ref{fig:exp_tv} 
by the person in the background in the upper left of the frame
is barely visible.

\vspace{-10pt}
\subsubsection{Ablation Studies}

\vspace{-5pt} \xhdr{Importance of Temporal Context.}
By directly comparing evaluation results of TRN with CNN and LSTM
baselines, we demonstrate the importance of explicitly modeling temporal
context  for online action detection. LSTMs capture long- and
short-term temporal patterns in the video  by
receiving accumulated
historical observations as inputs. Comparing TRN and LSTM measures the
benefit of incorporating predicted action features as future context.
CNN-based methods conduct online action detection by only considering
the image features at each time step. Simonyan \etal~\cite{simonyan2014two} build a
two-stream network and incorporate motion features
between adjacent video frames by using optical flow as inputs.
Table~\ref{table:th_detection} shows that this motion information
provides $1.5\%$ improvements. \textit{TRN-TS}
also takes optical flow as inputs and we can clearly see
a significant improvement ($83.7\%$ vs. $75.4\%$) on TVSeries.

\xhdr{Future Context: An ``Oracle'' Study.}  To
demonstrate the importance of using predictions of the future context,
we implemented an ``oracle'' baseline,
\textit{RNN-offline}. RNN-offline shares the same architecture as RNN
but uses the features extracted from both the current and future
frames as inputs. Note that RNN-offline uses future information and
thus is not an online model; the goal of this experiment is to show:
(1) the effectiveness of incorporating future information in action
detection, given access to future information
instead of predicting it; (2) and the performance gap between
the estimated future information of TRN and
the ``real'' future information of RNN-offline.
To permit a fair comparison, the input to RNN-offline is
the concatenation of the feature extracted from the current frame and
the average-pooled features of the next $\ell_d$
frames (where $\ell_d$ is the same as the number of
decoder steps of TRN).

\begin{table*}[t]
    \footnotesize
    \centering
    \resizebox{\textwidth}{!}{
        \begin{tabular}{@{\quad}l@{\;\;}@{\;\;}c@{\;\;}@{\;\;}c@{\;\;}@{\;\;}c@{\;\;}@{\;\;}c@{\;\;}@{\;\;}c@{\;\;}@{\;\;}c@{\;\;}@{\;\;}c@{\;\;}@{\;\;}c@{\;\;}@{\;\;}c@{\;\;}@{\;\;}c@{\;\;}@{\;\;}c@{\;\;}}
            \toprule
            & & \multicolumn{10}{c}{Portion of video} \\
            \cmidrule{3-12} 
            Method & Inputs & 0\%-10\% & 10\%-20\% & 20\%-30\% & 30\%-40\% & 40\%-50\% & 50\%-60\% & 60\%-70\% & 70\%-80\% & 80\%-90\% & 90\%-100\% \\
            \midrule
            {CNN~\cite{de2016online}} & \multirow{3}{*}{VGG} &61.0 &61.0 &61.2 &61.1 &61.2 &61.2 &61.3 &61.5 &61.4 &61.5 \\
            {LSTM~\cite{de2016online}} & &63.3 &64.5 &64.5 &64.3 &65.0 &64.7 &64.4 &64.3 &64.4 &64.3  \\
            \textbf{TRN} & &73.9 &74.3 &74.7 &74.7 &75.1 &75.1 &75.3 &75.2 &75.2 &75.3  \\
            \midrule
            {SVM~\cite{de2016online}} & FV &67.0 &68.4 &69.9 &71.3 &73.0 &74.0 &75.0 &76.4 &76.5 &76.8 \\
            \midrule
            \textbf{TRN} & TS &78.8 &79.6 &80.4 &81.0 &81.6 &81.9 &82.3 &82.7 &82.9 &83.3 \\
            \bottomrule
        \end{tabular}
        }
        \vspace{-5pt}
        \caption{
            \textit{Online action detection results 
            when only portions of action sequences are considered,}
            in terms of cAP (\%). 
            For example, 20\%-30\%  means that the first 30\% of frames
            of the action were seen and classified, but only frames
            in the 20\%-30\% time range
            were used to compute cAP.
        }
        \label{table:early}
        \vspace{-5pt}
\end{table*}


The results of RNN-offline are $41.6\%$, $85.3\%$, $47.3\%$ on
HDD, TVSeries, and THUMOS'14 datasets, respectively. Comparing
RNN-offline with the RNN baseline, we can see that the ``ground-truth"
future information significantly improves detection performance. We
also observe that the performance of TRN and RNN-offline are
comparable, \textit{even though TRN uses predicted rather than actual
future information}. This may be because TRN improves its
representation during learning by jointly optimizing current
and future action
recognition, while RNN-offline does not. We also evaluated TRN against
ED-based networks, by observing that ED can also improve its
representation by
jointly conducting action detection and anticipation. Thus,
comparisons between TRN with ED and its advanced
version~\cite{gao2017red} measure how much benefit comes purely from
explicitly incorporating anticipated future information.

\begin{table}
    \footnotesize
    \centering
    \begin{subtable}[t]{\linewidth}
        \centering
        \begin{tabular}
            {l@{\;}@{\;}c@{\;}@{\;}c@{\;}@{\;}c@{\;}@{\;}c@{\;}@{\;}c@{\;}@{\;}c@{\;}@{\;}c@{\;}@{\;}c@{\;}@{\,}r@{\;\;\;}}
            \toprule
            & \multicolumn{8}{c}{Time predicted into the future (seconds)} & \\
            \cmidrule{2-9}            Method & 0.25s & 0.5s & 0.75s & 1.0s & 1.25s & 1.5s & 1.75s & 2.0s & \;\;\;\;Avg \\
            \midrule
            {ED~\cite{gao2017red}} & 78.5 & 78.0 & 76.3 & 74.6 & 73.7 & 72.7 & 71.7 & 71.0 & 74.5 \\
            {RED~\cite{gao2017red}} & 79.2 & 78.7 & 77.1 & 75.5 & 74.2 & 73.0 & 72.0 & 71.2 & 75.1 \\
            \textbf{TRN} & 79.9 & 78.4 & 77.1 & 75.9 & 74.9 & 73.9 & 73.0 & 72.3 &\textbf{75.7} \\
            \bottomrule
        \end{tabular}
        \vspace{-3pt}
        \caption{
            Results on TVSeries dataset in terms of \textit{cAP} (\%).
        }
        \vspace{3pt}
    \end{subtable}
    \begin{subtable}[t]{\linewidth}
        \centering
        \begin{tabular}
            {l@{\;}@{\;}c@{\;}@{\;}c@{\;}@{\;}c@{\;}@{\;}c@{\;}@{\;}c@{\;}@{\;}c@{\;}@{\;}c@{\;}@{\;}c@{\;}@{\,}r@{\;\;\;}}
            \toprule
            & \multicolumn{8}{c}{Time predicted into the future (seconds)} & \\
            \cmidrule{2-9}            Method & 0.25s & 0.5s & 0.75s & 1.0s & 1.25s & 1.5s & 1.75s & 2.0s & \;\;\;\;Avg \\
            \midrule
            {ED~\cite{gao2017red}} & 43.8 & 40.9 & 38.7 & 36.8 & 34.6 & 33.9 & 32.5 & 31.6 & 36.6 \\
            {RED~\cite{gao2017red}} & 45.3 & 42.1 & 39.6 & 37.5 & 35.8 & 34.4 & 33.2 & 32.1 & 37.5 \\
            \textbf{TRN} & 45.1 & 42.4 & 40.7 & 39.1 & 37.7 & 36.4 & 35.3 & 34.3 &\textbf{38.9} \\
            \bottomrule
        \end{tabular}
        \vspace{-3pt}
        \caption{
            Results on THUMOS'14 dataset in terms of \textit{mAP} (\%).
        }        
        \vspace{3pt}
    \end{subtable}
    \vspace{-10pt}
    \caption{
        Action anticipation results of TRN compared to state-of-the-art
        methods using two-stream features.
    }
    \vspace{-5pt} 
    \label{table:anticipation}
\end{table}

\begin{table}[t]
    \centering
    \footnotesize
    \begin{tabular}
        {@{\;\;}l@{\;\;}@{\;\;}l@{\;\;}@{\;\;}c@{\;\;}@{\;\;}c@{\;\;}@{\;\;}c@{\;\;}@{\;\;}c@{\;\;}}
        \toprule
        & & \multicolumn{4}{c}{Decoder steps ($\ell_d$)}\\
        \cmidrule(r{2\cmidrulekern}){3-6}
        Dataset & Task & 4 & 6 & 8 & 10 \\
        \midrule
        \multirow{2}{*}{HDD}
        & {Online Action Detection}   & 39.9 & 40.8 & 40.1 & 39.6 \\
        & {Action Anticipation}       & 34.3 & 32.2 & 28.8 & 25.4 \\
        \midrule
        \multirow{2}{*}{TVSeries}
        & {Online Action Detection}   & 83.5 & 83.4 & 83.7 & 83.5 \\
        & {Action Anticipation}       & 77.7 & 76.4 & 75.7 & 74.1 \\
        \midrule
        \multirow{2}{*}{THUMOS'14}
        & {Online Action Detection}   & 46.0 & 45.4 & 47.2 & 46.4 \\
        & {Action Anticipation}       & 42.6 & 39.4 & 38.9 & 35.0 \\
        \bottomrule

    \end{tabular}
    \vspace{-5pt}
    \caption{
        Online action detection and action anticipation results of TRN
        with decoder steps $\ell_d=4,6,8,10$.
    }
    \vspace{-12pt}
    \label{table:decoder}
\end{table}


\xhdr{Different Decoder Steps.} Finally, we evaluated the
effectiveness of different numbers of decoder steps $\ell_d$. In
particular, we tested $\ell_d=4,6,8,10$, as shown in
Table~\ref{table:decoder}, where the performance of action
anticipation is averaged over the decoder steps. The results show that
a larger number of decoder steps does not guarantee better
performance. This is because anticipation accuracy usually
decreases for longer future sequences, and
thus creates more  noise in the input features of STA.  To be
clear, we follow the state-of-the-art~\cite{gao2017red} and set
$\ell_d$ to 2 video seconds (6 frames in HDD,  8 frames in
TVSeries and THUMOS'14) when comparing
with baseline methods of online action detection in
Tables~\ref{table:hdd_detection},~\ref{table:tv_detection},
and~\ref{table:th_detection}.

\subsubsection{Evaluation of Different Stages of Action.}

We evaluated TRN when only a fraction of each action is
considered, and compared with published
results~\cite{de2016online} on TVSeries. Table~\ref{table:early}
shows that TRN significantly outperforms existing methods at every
time stage. Specifically, when we compare \textit{TRN-TS} with the
best baseline \textit{SVM-FV}, the performance gaps between these
two methods are roughly in ascending order as less and less of the
actions are observed
(the gaps are $6.5\%$, $6.4\%$, $6.3\%$, $7.3\%$, $7.9\%$, $8.6\%$,
$9.7\%$, $10.5\%$, $11.2\%$ and $11.8\%$ from actions at $100\%$
observed to those are $10\%$ observed). This indicates the advantage of
our approach at earlier stages of actions.

\subsubsection{Evaluation of Action Anticipation.}

We also evaluated TRN on predicting actions for up to 2 seconds into
the future, and  compare our approach with the
state-of-the-art~\cite{gao2017red} in
Table~\ref{table:anticipation}. The results show that TRN
performs better than RED and ED baselines (mcAP of
$75.7\%$ vs.~$75.1\%$ vs.~$74.5\%$ on TVSeries and mAP of $38.9\%$
vs.~$37.5\%$ vs.~$36.6\%$ on THUMOS'14). The average of anticipation
results over the next $2$ seconds on HDD dataset is $32.2\%$ in terms
of per-frame mAP.

\section{Conclusion}

In this paper, we propose Temporal Recurrent Network (TRN) to model
greater temporal context and evaluate on the online action detection
problem. Unlike previous methods that consider only historical
temporal consistencies, TRN jointly models the historical and future
temporal context under the constraint of the online setting. Experimental
results on three popular datasets demonstrate that incorporating
predicted future information improves learned representation of actions
and significantly outperforms the state-of-the-art. Moreover, TRN
shows more advantage at earlier stages of actions, and in predicting future
actions. More generally, we believe that our approach of incorporating estimated future
information could benefit many other online tasks, such as
video object localization and tracking, and plan to pursue this in future work.

\section{Acknowledgments}

This work was supported in part by the Intelligence
Advanced Research Projects Activity (IARPA) via Department of Interior/Interior Business Center (DOI/IBC)
contract number D17PC00345, the National Science Foundation (CAREER IIS-1253549), the IU Office of the Vice
Provost for Research, the College of Arts and Sciences,
the School of Informatics, Computing, and Engineering
through the Emerging Areas of Research Project ``Learning:
Brains, Machines, and Children,'' and Honda Research
Institute USA. The views and conclusions contained in
this paper are those of the authors and should not be
interpreted as representing the official policies,
either expressly or implied, of the U.S. Government, or any sponsor.

{\small
\bibliographystyle{ieee}
\bibliography{egbib}
}

\end{document}